\documentclass[accepted]{uai2026} 

\usepackage[american]{babel}
\hypersetup{colorlinks=false, pdfborder={0 0 0}}

\usepackage{subcaption}

\usepackage{natbib} 
\usepackage{mathtools} 
\usepackage{booktabs} 
\usepackage{tikz} 
\usetikzlibrary{arrows.meta, backgrounds}
\usepackage{bm} 
\usepackage{amssymb} 

\usepackage{xcolor}

\title{Hypernetwork-based approach for grid-independent functional data clustering}

\author[1]{Anirudh~Thatipelli}
\author[1]{Ali~Siahkoohi}
\affil[1]{%
    Department of Computer Science\\
    University of Central Florida
}
  
\begin{document}
\maketitle

\begin{abstract}
Functional data clustering is concerned with grouping functions that share similar structure, yet most existing methods implicitly operate on sampled grids, causing cluster assignments to depend on resolution, sampling density, or preprocessing choices rather than on the underlying functions themselves. To address this limitation, we introduce a framework that maps discretized function observations---at arbitrary resolution and on arbitrary grids---into a fixed-dimensional vector space via an auto-encoding architecture. The encoder is a hypernetwork that maps coordinate-value pairs to the weight space of an implicit neural representation (INR), which serves as the decoder. Because INRs represent functions with very few parameters, this design yields compact representations that are decoupled from the sampling grid, while the hypernetwork amortizes weight prediction across the dataset. Clustering is then performed in this weight space using standard algorithms, making the approach agnostic to both the discretization and the choice of clustering method. By means of synthetic and real-world experiments in high-dimensional settings, we demonstrate competitive clustering performance that is robust to changes in sampling resolution---including generalization to resolutions not seen during training.
\end{abstract}

\definecolor{cblue}{HTML}{4E79A7}
\definecolor{cred}{HTML}{E15759}
\definecolor{cgray}{HTML}{6B7280}
\definecolor{cbg}{HTML}{F5F6F8}

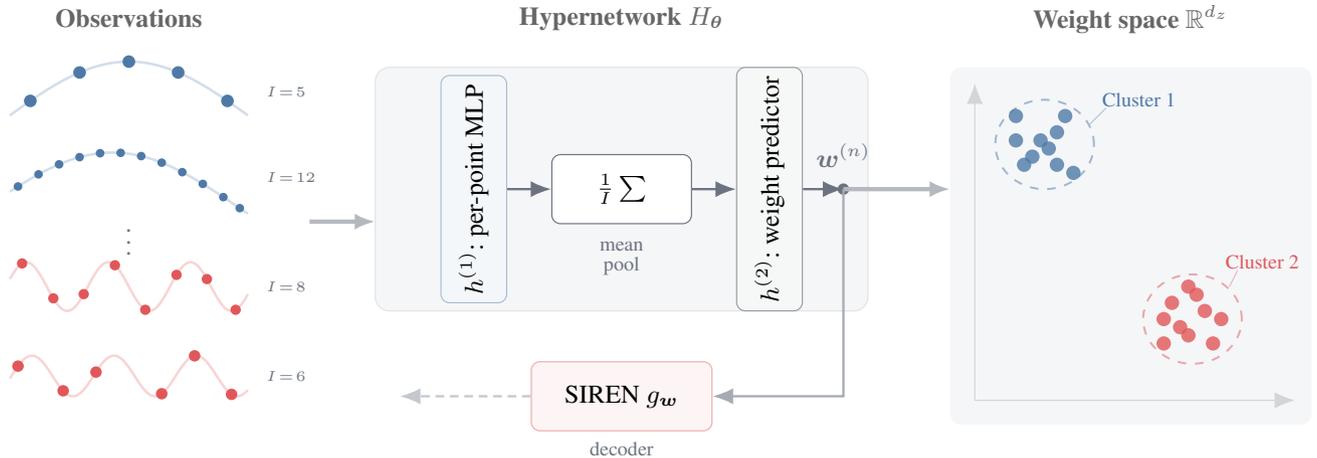
\begin{figure*}[t]
\centering
\resizebox{\textwidth}{!}{%
\begin{tikzpicture}[
  >={Latex[length=2.5mm, width=2mm]},
  every node/.style={font=\small},
  block/.style={
    draw=cgray, fill=white, rounded corners=3pt,
    minimum height=0.85cm, minimum width=1.7cm,
    font=\small, inner sep=4pt
  },
  pipelinearrow/.style={->, thick, cgray},
  bigarrow/.style={->, line width=1.5pt, cgray!50},
  stagelabel/.style={font=\small\bfseries, text=black!60},
  sublabel/.style={font=\scriptsize, text=cgray},
]

\node[stagelabel] at (1.5, 5.3) {Observations};

\begin{scope}[shift={(0, 3.95)}]
  \draw[cblue!25, line width=0.9pt, smooth] plot[domain=0.05:2.95, samples=60]
    (\x, {0.12 + 0.7*sin(\x*60)});
  \foreach \x in {0.3, 0.9, 1.5, 2.1, 2.7}
    \fill[cblue] (\x, {0.12 + 0.7*sin(\x*60)}) circle (2.2pt);
  \node[font=\tiny, cgray, anchor=west] at (3.05, 0.45) {$I\!=\!5$};
\end{scope}

\begin{scope}[shift={(0, 2.9)}]
  \draw[cblue!25, line width=0.9pt, smooth] plot[domain=0.05:2.95, samples=60]
    (\x, {0.1 + 0.65*sin(\x*60 + 12)});
  \foreach \x in {0.15,0.40,0.65,0.90,1.15,1.40,1.65,1.90,2.15,2.40,2.65,2.85}
    \fill[cblue] (\x, {0.1 + 0.65*sin(\x*60 + 12)}) circle (1.5pt);
  \node[font=\tiny, cgray, anchor=west] at (3.05, 0.45) {$I\!=\!12$};
\end{scope}

\node[cgray] at (1.5, 2.65) {$\vdots$};

\begin{scope}[shift={(0, 1.55)}]
  \draw[cred!25, line width=0.9pt, smooth] plot[domain=0.05:2.95, samples=80]
    (\x, {0.45 + 0.3*sin(\x*360)});
  \foreach \x in {0.2, 0.58, 0.95, 1.33, 1.70, 2.08, 2.45, 2.80}
    \fill[cred] (\x, {0.45 + 0.3*sin(\x*360)}) circle (1.8pt);
  \node[font=\tiny, cgray, anchor=west] at (3.05, 0.45) {$I\!=\!8$};
\end{scope}

\begin{scope}[shift={(0, 0.45)}]
  \draw[cred!25, line width=0.9pt, smooth] plot[domain=0.05:2.95, samples=80]
    (\x, {0.45 + 0.25*sin(\x*360 - 25)});
  \foreach \x in {0.15, 0.7, 1.1, 1.9, 2.3, 2.75}
    \fill[cred] (\x, {0.45 + 0.25*sin(\x*360 - 25)}) circle (2.0pt);
  \node[font=\tiny, cgray, anchor=west] at (3.05, 0.45) {$I\!=\!6$};
\end{scope}

\draw[bigarrow] (3.7, 2.8) -- (4.5, 2.8);

\node[stagelabel] at (7.5, 5.3) {Hypernetwork $H_{\bm{\theta}}$};

\begin{scope}[on background layer]
  \fill[cbg, rounded corners=5pt] (4.5, 1.7) rectangle (10.5, 4.7);
  \draw[cgray!20, rounded corners=5pt] (4.5, 1.7) rectangle (10.5, 4.7);
\end{scope}

\node[draw=cblue!50, fill=cblue!6, rounded corners=3pt,
      minimum height=2.2cm, minimum width=0.8cm, inner sep=3pt]
  (h1) at (5.7, 3.2) {\rotatebox{90}{\footnotesize $h^{(1)}$: per-point MLP}};

\node[block] (pool) at (7.5, 3.2) {$\tfrac{1}{I}\sum$};
\node[sublabel, align=center] at (7.5, 2.35) {mean\\[-1pt]pool};

\node[draw=cgray!70, fill=cgray!6, rounded corners=3pt,
      minimum height=2.2cm, minimum width=0.8cm, inner sep=3pt]
  (h2) at (9.3, 3.2) {\rotatebox{90}{\footnotesize $h^{(2)}$: weight predictor}};

\draw[pipelinearrow] (h1.east) -- (pool);
\draw[pipelinearrow] (pool) -- (h2.west);

\node[block, draw=cred!50, fill=cred!6, minimum width=2.2cm] (siren) at (7.5, 0.65)
  {SIREN $g_{\bm{w}}$};
\node[sublabel] at (7.5, 0.0) {decoder};

\coordinate (branch) at (10.2, 3.2);
\draw[pipelinearrow] (h2.east) -- (branch);
\fill[cgray] (branch) circle (2pt);
\node[font=\small, cgray, above] at (10.2, 3.35) {$\bm{w}^{(n)}$};

\draw[bigarrow] (branch) -- (11.5, 3.2);

\draw[pipelinearrow, cgray!60] (branch) -- (10.2, 0.65) -- (siren.east);

\draw[pipelinearrow, densely dashed, cgray!40]
  (siren.west) -- (4.8, 0.65);

\node[stagelabel] at (13.7, 5.3) {Weight space $\mathbb{R}^{d_z}$};

\begin{scope}[on background layer]
  \fill[cbg, rounded corners=4pt] (11.5, 0.3) rectangle (15.9, 4.7);
\end{scope}

\draw[cgray!30, thin, ->] (11.8, 0.6) -- (11.8, 4.5);
\draw[cgray!30, thin, ->] (11.8, 0.6) -- (15.7, 0.6);

\foreach \px/\py in {
  12.5/3.6, 12.8/3.9, 12.3/4.1, 13.0/3.4,
  12.6/3.8, 12.4/3.5, 12.9/4.1, 12.7/3.7,
  12.3/3.8, 12.8/3.5}
  \fill[cblue, opacity=0.8] (\px, \py) circle (2.5pt);

\foreach \px/\py in {
  14.3/1.5, 14.6/1.7, 14.1/1.3, 14.5/1.9,
  14.4/1.4, 14.8/1.6, 14.2/1.8, 14.7/1.3,
  14.4/2.0, 14.1/1.6}
  \fill[cred, opacity=0.8] (\px, \py) circle (2.5pt);

\draw[cblue, dashed, line width=0.7pt, opacity=0.5] (12.65, 3.75) ellipse (0.6 and 0.55);
\draw[cred, dashed, line width=0.7pt, opacity=0.5] (14.45, 1.6) ellipse (0.6 and 0.55);

\node[font=\scriptsize, cblue] at (13.8, 4.3) {Cluster 1};
\draw[cblue, thin, opacity=0.4] (13.4, 4.2) -- (13.2, 4.0);
\node[font=\scriptsize, cred] at (15.3, 2.3) {Cluster 2};
\draw[cred, thin, opacity=0.4] (15.0, 2.2) -- (14.9, 2.0);

\end{tikzpicture}%
}
\caption{Overview of the proposed framework. Discretized function observations---sampled at varying resolutions and on potentially different grids (left)---are mapped by the hypernetwork encoder $H_{\bm{\theta}}$ into the weight space of a SIREN decoder (center). The per-point network $h^{(1)}$ processes each coordinate-value pair independently, mean pooling aggregates features into a fixed-dimensional vector, and the weight predictor $h^{(2)}$ outputs the SIREN weights $\bm{w}$. The resulting weight vectors lie in a common Euclidean space regardless of the input discretization, enabling direct application of standard clustering algorithms (right). The dashed arrow denotes the reconstruction loss (equation~\eqref{eq:loss}) used to train the hypernetwork end-to-end. Code to reproduce the results presented in this work is available at \href{https://github.com/luqigroup/hypercluster}{\textcolor{blue}{github.com/luqigroup/hypercluster}}.}
\label{fig:method}
\end{figure*}

\section{Introduction}\label{sec:intro}

Functional data analysis is concerned with data samples that are best understood as functions---e.g., time series, spatial fields, or spatially varying material properties---rather than finite-dimensional vectors \citep{ramsay2005functional, ferraty2006nonparametric}. In many scientific and engineering applications, each observation represents a continuous process that is sampled on a finite grid of points, and the goal is to identify groups of functions that share similar structure. This task, known as functional data clustering, arises in fields as diverse as medical imaging, climate science, and geophysics \citep{seydoux2020clustering, zhang2023functional, siahkoohi2026multiscale}. A fundamental challenge is that the discretization used to observe each function is an artifact of the measurement process, not an intrinsic property of the underlying object. When cluster assignments change with the choice of grid, practitioners face an undesirable source of uncertainty: downstream conclusions depend on a nuisance parameter---the discretization---rather than on the data alone. Cluster assignments should therefore be invariant to the choice of grid, yet most existing methods implicitly depend on it.

Existing approaches either operate directly in infinite-dimensional function spaces---requiring intricate algorithm design that cannot leverage standard finite-dimensional clustering tools---or first project functions onto basis expansions (e.g., B-splines, functional principal component analysis (PCA)) whose coefficients are then clustered \citep{abraham2003bspline, jacques2013funclust, zhang2023functional}. The latter two-stage strategy is appealing because it reduces the problem to standard finite-dimensional clustering, but classical instantiations have important limitations. First, they require all functions to be observed on a common grid, often necessitating a smoothing or interpolation preprocessing step; even methods designed for sparse or irregular observations \citep{james2003clustering, yao2005functional} remain tied to a fixed basis system. Second, \citet{tarpey2003clustering} show that applying $K$-means~\citep{lloyd1982kmeans} to basis-expansion coefficients is not equivalent to clustering in the functional $L^2$ space unless the basis is orthogonal---so the choice of basis can itself distort cluster structure. Third, basis-expansion methods scale poorly to functions whose input domain $\Omega$ is higher-dimensional---e.g., images or spatial fields where $\Omega \subset \mathbb{R}^2$ or $\mathbb{R}^3$---since the number of basis functions grows exponentially with the dimension of $\Omega$ \citep{ramsay2005functional}. The most closely related work, FAEclust \citep{singh2025faeclust}, takes a step toward learning-based functional data clustering by jointly training a functional autoencoder with a clustering objective. While FAEclust handles vector-valued functions---i.e., functions with multi-dimensional output---its encoder and decoder are defined via integral operators over a compact interval $\mathcal{T} \subset \mathbb{R}$, restricting the function \emph{input domain} to a single variable (e.g., time). It further inherits the remaining limitations: a fixed common grid, and tight coupling between the learned representation and a specific clustering algorithm---changing the clustering method requires retraining the entire model.

In this work, we propose a hypernetwork-based framework that addresses these limitations (Figure~\ref{fig:method}). The core idea is to map each function---observed at arbitrary resolution and on an arbitrary grid---into the weight space of an implicit neural representation (INR) \citep{sitzmann2020siren}. A hypernetwork encoder \citep{ha2017hypernetworks, mayer2024ple} processes coordinate-value pairs through a mesh-independent aggregation mechanism \citep{stuart2025autoencoders} and outputs the weights of an INR decoder that can reconstruct the function at any coordinate. Because the resulting weight vectors are compact and decoupled from the sampling grid, clustering can be performed directly in this space using off-the-shelf algorithms---e.g., $K$-means---making the pipeline agnostic to both the discretization and the clustering method. Following \citet{sakarvadia2026false}, we train on data observed at multiple resolutions simultaneously rather than relying on zero-shot generalization, and demonstrate robust generalization to held-out resolutions at inference time.

The main contributions of this work are as follows: (i) we propose a hypernetwork-based framework for functional data clustering that is, by construction, independent of the discretization grid---enabling robust clustering on resolutions and grid geometries not encountered during training; (ii) because the decoder is an INR that takes coordinates in $\mathbb{R}^d$ as input, our approach scales naturally to functions defined on higher-dimensional input domains---e.g., images ($\Omega \subset \mathbb{R}^2$) or volumetric fields ($\Omega \subset \mathbb{R}^3$)---whereas existing functional autoencoder methods, including FAEclust, are restricted to functions of a single input variable ($\Omega \subset \mathbb{R}$, e.g., time series) despite supporting multi-dimensional output; (iii) by decoupling representation learning from clustering, our framework allows practitioners to apply standard clustering algorithms directly in the learned weight space, without requiring infinite-dimensional or task-specific adaptations; and (iv) the framework natively handles heterogeneous grids---i.e., functions observed at different resolutions and on different point sets within the same dataset---without any preprocessing or smoothing step.

In the following sections, we first formalize the problem setup and describe each component of the proposed framework---the mesh-independent encoder, the INR-based decoder, and the hypernetwork tying them together (Section~\ref{sec:method}). We then evaluate on synthetic and real-world datasets, demonstrating competitive clustering performance across a range of resolutions and grid geometries (Section~\ref{sec:exps}), before discussing related work (Section~\ref{sec:related}) and limitations and future directions (Section~\ref{sec:discussion}).

\section{Method}
\label{sec:method}

The goal of the proposed framework is to map each function into a fixed-dimensional vector space in a manner that is (i) independent of the discretization, (ii) information-preserving, and (iii) as compact as possible. To this end, we adopt an autoencoding architecture: an \emph{encoder} mapping from function space to a vector space, and a \emph{decoder} mapping back. For the encoder, we follow \citet{stuart2025autoencoders}, who provide a principled architecture for mapping discretized function observations to a finite-dimensional latent space invariant to the number and placement of discretization points. For the decoder, we treat the latent vectors as \emph{weights of an INR}---a small network that reconstructs the function at given spatial coordinates. This is motivated by two observations: (i) INR architectures such as SIREN \citep{sitzmann2020siren} very accurately approximate target functions; and (ii) INRs are inherently compact, since a major application is data compression---representing a function entirely via the weights of a small network \citep{dupont2021coin, roddenberry2024inr}.

\subsection{Setup and notation}
\label{sec:problem-def}

We are concerned with clustering a collection of functions that may be observed on different grids. Formally, let $(\mathcal{U}, \|\cdot\|)$ be a separable Banach space of functions with domain $\Omega \subseteq \mathbb{R}^d$ and range $\mathbb{R}^m$, and consider a collection of $N$ functions $\{u^{(n)}\}_{n=1}^N \subset \mathcal{U}$, each observed at a potentially different set of locations $\{\bm{x}_i^{(n)}\}_{i=1}^{I_n} \subset \Omega$, yielding
\begin{equation}
    \bm{u}^{(n)} = \left\{ \left(\bm{x}_i^{(n)},\, u^{(n)}(\bm{x}_i^{(n)})\right) \right\}_{i=1}^{I_n} \subset \Omega \times \mathbb{R}^m.
    \label{eq:observations}
\end{equation}
In the above expression, each observation pairs a spatial coordinate $\bm{x}_i^{(n)} \in \Omega$ with the corresponding function value $u^{(n)}(\bm{x}_i^{(n)}) \in \mathbb{R}^m$. Both the number of observation points $I_n$ and their locations may differ across functions---for instance, one function might be sampled on a uniform grid of $16$ points while another is observed at $100$ irregularly spaced locations. This heterogeneity poses a fundamental challenge for methods that operate directly on sampled values, since two discretizations of the same function would be treated as different objects---conflating measurement variability with genuine differences between functions. The goal is to assign each function to one of $K$ clusters such that the assignments are invariant to the discretization grid. To this end, we seek a mapping $\mathcal{E}_{\bm{\theta}} \colon \mathcal{U} \to \mathcal{Z} = \mathbb{R}^{d_z}$ that produces a fixed-dimensional representation $\bm{w}^{(n)} = \mathcal{E}_{\bm{\theta}}(\bm{u}^{(n)}) \in \mathbb{R}^{d_z}$ for each function. We describe the construction of this mapping in the next three subsections.

\subsection{Encoding discretized functions}
\label{sec:encoder}

A standard neural network encoder expects a fixed-size input vector, which implicitly ties the representation to a particular grid. To address this limitation, we follow \citet{stuart2025autoencoders} and define the encoding map as
\begin{equation}
    \bm{f}(u;\, \bm{\theta}) = \rho\!\left(\int_{\Omega} \kappa\!\left(\bm{x},\, u(\bm{x});\, \bm{\theta}\right) \mathrm{d}\bm{x};\, \bm{\theta}\right) \in \mathbb{R}^{d_z},
    \label{eq:encoder}
\end{equation}
where $\kappa \colon \Omega \times \mathbb{R}^m \times \Theta \to \mathbb{R}^{\ell}$ is a neural network that processes coordinate-value pairs $(\bm{x}, u(\bm{x}))$, and $\rho \colon \mathbb{R}^{\ell} \times \Theta \to \mathbb{R}^{d_z}$ is a subsequent mapping layer. In the above expression, $\kappa$ acts pointwise---taking a single coordinate and its associated function value and producing a feature vector of dimension $\ell$---while $\rho$ maps the aggregated features to the latent dimension $d_z$. The key insight is that the integral over $\Omega$ aggregates information from the entire domain in a way that does not depend on any particular grid.

In practice, we do not have access to the continuous function $u$ but only to the discrete observations $\bm{u}$ in equation~\eqref{eq:observations}. The integral in equation~\eqref{eq:encoder} is therefore approximated by a mean over the available points:
\begin{equation}
    \bm{f}(\bm{u};\, \bm{\theta}) = \rho\!\left(\frac{1}{I} \sum_{i=1}^{I} \kappa\!\left(\bm{x}_i,\, u(\bm{x}_i);\, \bm{\theta}\right);\, \bm{\theta}\right).
    \label{eq:encoder-discrete}
\end{equation}
This mean-pooling aggregation has two important consequences. First, it renders the encoder \emph{permutation invariant}~\citep{zaheer2017deep}: the output does not depend on the ordering of the input points, which is natural since they are unordered samples from a function. Second, it makes the encoder \emph{mesh-independent}: because the normalized sum approximates the same underlying integral regardless of discretization, different samplings of the same function produce similar outputs. \citet{stuart2025autoencoders} show that this architecture satisfies compatibility conditions ensuring well-definedness in function space---i.e., the encoder output converges to a well-defined limit as the discretization is refined. The encoder in equation~\eqref{eq:encoder-discrete} maps observations to a fixed-dimensional feature vector, but we have not yet specified how this feature vector is decoded back into a function. In the next subsection, we describe the decoder architecture that enables this reconstruction.

\subsection{Functions as neural network weights}
\label{sec:inr}

Rather than storing function values on a grid, an INR parameterizes a function as a small neural network $g_{\bm{w}} \colon \mathbb{R}^d \to \mathbb{R}^m$ whose weights $\bm{w}$ fully determine the function \citep{sitzmann2020siren}. Given the weights, the function can be evaluated at \emph{any} coordinate $\bm{x} \in \Omega$ via a forward pass through $g_{\bm{w}}$, decoupling the representation from any particular grid. This makes INRs a natural fit for our framework: a single weight vector $\bm{w}$ provides a compact, grid-free description of the underlying function.

For the decoder, we adopt the SIREN architecture \citep{sitzmann2020siren}, which uses sinusoidal activations. Each hidden layer is defined as
\begin{equation}
    \bm{\phi}_l(\bm{h}) = \sin\!\left(\omega_0 \left(\bm{W}_l \bm{h} + \bm{b}_l\right)\right), \quad l = 1, \ldots, L-1,
    \label{eq:siren-layer}
\end{equation}
where $\bm{W}_l$ and $\bm{b}_l$ are the weight matrix and bias of the $l$-th layer, and $\omega_0$ is a frequency hyperparameter controlling the bandwidth of the representation. In the above expression, $\bm{h}$ denotes the input to the layer---i.e., the output of the previous layer, or the spatial coordinate $\bm{x}$ for $l = 1$. Intuitively, each layer applies a learned affine transformation followed by a sinusoidal nonlinearity, allowing the network to build up increasingly complex signal representations through compositions of periodic functions. The final layer applies a linear transformation without the activation, and the full network reads
\begin{equation}
    g_{\bm{w}}(\bm{x}) = \bm{W}_L \bm{\phi}_{L-1} \circ \cdots \circ \bm{\phi}_1(\bm{x}) + \bm{b}_L,
    \label{eq:siren-full}
\end{equation}
where $\bm{w} = \{\bm{W}_l, \bm{b}_l\}_{l=1}^L$ denotes the collection of all weights and biases. Sinusoidal activations enable accurate representation of signals and their derivatives \citep{sitzmann2020siren, roddenberry2024inr}, and SIREN comes with a well-studied initialization scheme that ensures stable training. From the perspective of clustering, INRs offer two properties that are difficult to achieve with grid-based representations. The first is \emph{compactness}: a small SIREN---with only a few hidden layers and modest width---can represent complex functions using far fewer parameters than an equivalent grid-based representation, yielding a low-dimensional weight vector well-suited for downstream analysis. The second is \emph{dimension agnosticism}: since the INR takes coordinates $\bm{x} \in \mathbb{R}^d$ as input, the same architecture applies to functions defined on input domains of any dimension---from time series ($\Omega \subset \mathbb{R}$) to images ($\Omega \subset \mathbb{R}^2$) to volumetric fields ($\Omega \subset \mathbb{R}^3$)---without any modification.

Having defined both the encoder and the decoder, we now describe how they are combined into a single end-to-end architecture via a hypernetwork.

\subsection{Amortized weight prediction via hypernetworks}
\label{sec:hypernetwork}

One could fit a separate SIREN to each function independently, but this requires solving a per-function optimization problem and does not share information across the dataset. A hypernetwork \emph{amortizes} this process: it is a neural network that takes data as input and directly outputs the weights of another network \citep{ha2017hypernetworks}. In our framework, the hypernetwork learns a single feedforward mapping from discretized function observations to SIREN weights, generalizing to new functions at inference time without any per-function optimization.

\paragraph{Architecture.} Our hypernetwork $H_{\bm{\theta}}$ combines the mesh-independent encoding layer of \citet{stuart2025autoencoders} with a weight-prediction mechanism following the design principles of \citet{mayer2024ple}. It maps the discretized observations $\bm{u}^{(n)}$ to the SIREN weight vector $\bm{w}^{(n)}$ in two stages:
\begin{equation}
    \bm{w}^{(n)} = H_{\bm{\theta}}\!\left(\bm{u}^{(n)}\right) = h^{(2)}\!\left(\frac{1}{I_n} \sum_{i=1}^{I_n} h^{(1)}\!\left(\bm{x}_i^{(n)},\, u^{(n)}(\bm{x}_i^{(n)})\right)\right).
    \label{eq:hypernetwork}
\end{equation}
In the first stage, $h^{(1)} \colon \Omega \times \mathbb{R}^m \to \mathbb{R}^{\ell}$ independently processes each coordinate-value pair into a feature vector of dimension $\ell$. These per-point features are then aggregated via mean pooling, yielding a single intermediate representation that summarizes the function regardless of how many points were observed. In the second stage, $h^{(2)} \colon \mathbb{R}^{\ell} \to \mathbb{R}^{d_z}$ maps this intermediate representation to the full weight vector of the target SIREN. Here, $h^{(1)}$ corresponds to the $\kappa$ network in equation~\eqref{eq:encoder-discrete}, while $h^{(2)}$ extends $\rho$ to predict all SIREN weights from the pooled features.

This two-stage design inherits the permutation invariance and mesh-independence properties established in Section~\ref{sec:encoder}, while the modular structure of $h^{(2)}$---which predicts the weights of each SIREN layer through an independent head \citep{mayer2024ple}---keeps the parameter count manageable and allows per-layer predictions to be computed in parallel. A practical advantage is that the hypernetwork automatically constructs one prediction head per SIREN layer, so changing the decoder architecture requires no manual adjustment---the only user-specified hyperparameters are the hidden-layer widths of $h^{(1)}$ and $h^{(2)}$.

\paragraph{Training objective.} We train the hypernetwork end-to-end by minimizing the mean squared reconstruction error over the training set $\{\bm{u}^{(n)}\}_{n=1}^N$:
\begin{equation}
    \mathcal{L}(\bm{\theta}) = \frac{1}{N} \sum_{n=1}^{N} \frac{1}{I_n} \sum_{i=1}^{I_n} \left\| u^{(n)}(\bm{x}_i^{(n)}) - g_{\bm{w}^{(n)}}(\bm{x}_i^{(n)}) \right\|_2^2,
    \label{eq:loss}
\end{equation}
where $\bm{w}^{(n)} = H_{\bm{\theta}}(\bm{u}^{(n)})$ are the predicted SIREN weights for the $n$-th function and $g_{\bm{w}^{(n)}}$ is the corresponding SIREN decoder. In the above expression, the SIREN is evaluated at the same coordinates where the function was observed, and the normalization by $I_n$ ensures that functions with more observation points do not dominate the loss. The entire pipeline---from observations through the hypernetwork to the SIREN evaluation---is differentiable, enabling end-to-end training via standard gradient-based optimization. It is worth noting that the loss in equation~\eqref{eq:loss} is purely a reconstruction objective, with no clustering-specific term. This is a deliberate design choice: by decoupling representation learning from clustering, the learned weight space can be reused with any clustering algorithm---$K$-means, spectral clustering \citep{vonluxburg2007spectral}, Gaussian mixture models (GMMs)---without retraining.

While our encoder is mesh-independent by construction, we do not rely on zero-shot generalization to unseen resolutions at inference time. Recent work has demonstrated that machine-learned operators---even those with theoretical discretization-invariance---fail to generalize reliably when evaluated at resolutions different from those seen during training, effectively encountering a distribution shift \citep{sakarvadia2026false}. Following the multi-resolution training protocol advocated by \citet{sakarvadia2026false}, we instead train the hypernetwork on data observed at multiple resolutions simultaneously---e.g., grids of size $\{16, 32, 64\}$. This ensures that the learned mapping has seen a range of discretizations during training, and we find that it then generalizes robustly to held-out resolutions (e.g., $\{24, 48, 96, 128, 256\}$) at inference time.

\paragraph{Clustering in weight space.} Once trained, the hypernetwork produces a weight vector $\bm{w}^{(n)} \in \mathbb{R}^{d_z}$ for each function via a single forward pass. Because both the hypernetwork and the SIREN evaluation are fully batched, an entire dataset of functions can be encoded and reconstructed in parallel on a GPU, making the embedding step efficient even for large collections. These vectors live in a common Euclidean space regardless of the discretization used for each function, and standard clustering algorithms---e.g., $K$-means---can be applied directly to $\{\bm{w}^{(n)}\}_{n=1}^N$. Functions that share the same underlying shape are mapped to nearby points in weight space, even if they are observed on different grids, so that cluster assignments reflect functional similarity rather than artifacts of the observation process.

\section{Experiments}
\label{sec:exps}

The purpose of the following experiments is to evaluate whether mapping functions into INR weight space yields cluster assignments that are invariant to the discretization grid. To this end, we assess robustness to (i) sampling resolution and (ii) the choice of clustering algorithm across three families of functional data that span both multi-dimensional domains and multi-dimensional ranges: discretized image data (MNIST; $\Omega \subset \mathbb{R}^2$, scalar-valued), medical images (Kvasir~\cite{pogorelov2017kvasir}; $\Omega \subset \mathbb{R}^2$, $\mathbb{R}^3$-valued), and multivariate time series from the UEA archive (ERing~\cite{bagnall2018uea}; $\Omega \subset \mathbb{R}$, $\mathbb{R}^{4}$-valued). In all experiments, the representation is trained once and clustering is evaluated across multiple discretizations---including resolutions not seen during training---without retraining. Clustering performance is measured by Adjusted Mutual Information (AMI)~\citep{vinh2010ami} and Adjusted Rand Index (ARI)~\citep{hubert1985ari}, both of which are normalized so that a random partition scores zero in expectation and perfect agreement scores one. AMI quantifies the shared information between predicted and ground-truth label assignments, while ARI measures the fraction of sample pairs that are consistently grouped or separated across the two partitions. Uniform Manifold Approximation and Projection (UMAP)~\cite{McInnes2018} is used for visualization. Architectural details and training protocols are provided in Appendix~\ref{appendix:impl_details}.

\subsection{MNIST Dataset}
\label{subsec:mnist_example}

\begin{figure}[t]
    \centering
    \setlength{\tabcolsep}{3pt}
    \renewcommand{\arraystretch}{1.1}

    \begin{tabular}{c *{3}{c}}
        & \textbf{Cluster 0} & \textbf{Cluster 1} & \textbf{Cluster 3} \\[0.4em]

        \rotatebox{90}{\large $14\times 14$} &
        \includegraphics[width=0.22\linewidth]{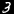} &
        \includegraphics[width=0.22\linewidth]{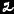} &
        \includegraphics[width=0.22\linewidth]{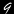} \\[0.6em]

        \rotatebox{90}{\large $28\times 28$} &
        \includegraphics[width=0.22\linewidth]{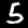} &
        \includegraphics[width=0.22\linewidth]{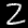} &
        \includegraphics[width=0.22\linewidth]{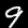} \\[0.6em]

        \rotatebox{90}{\large $56\times 56$} &
        \includegraphics[width=0.22\linewidth]{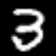} &
        \includegraphics[width=0.22\linewidth]{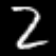} &
        \includegraphics[width=0.22\linewidth]{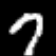} \\
    \end{tabular}

    \caption{Representative MNIST samples arranged in a grid where rows correspond to discretization resolutions and columns correspond to learned clusters. Digits are clustered according to their semantic meaning, not by their discretization resolution.}
    \label{fig:mnist_cluster_samples}
    \vspace{-1.5em}
\end{figure}

MNIST~\citep{lecun1998mnist} serves as a controlled setting where the ground-truth cluster structure---i.e., digit identity---is known. Each digit image is treated as a function $f: [0, 1]^2 \rightarrow [0, 1]$, where coordinates correspond to normalized pixel locations and function values denote intensity. Discretizations are constructed at resolutions $r \in \{14, 28, 56\}$ via bilinear interpolation, and each discretized image is represented as an unordered set of coordinate-intensity pairs. The SIREN decoder has $d_z = 81$ parameters. Training uses multi-resolution sampling over $\mathrm{R_{train}} = \{14, 28, 56\}$, and evaluation includes held-out resolutions $\mathrm{R_{test}} = \{7, 112\}$. Figure~\ref{fig:mnist_cluster_samples} shows that the learned clusters are discretization-invariant.

Table~\ref{tab:mnist_eval_diff_resolns} reports clustering performance. As a reference, applying $K$-means directly to flattened pixel intensities yields AMI~$\approx 0.42$ and ARI~$\approx 0.30$, confirming that the learned weight-space representation provides substantially richer structure for clustering. For seen resolutions ($14$, $28$, $56$), ARI and AMI remain statistically consistent, confirming that multi-resolution training prevents overfitting to any particular discretization. As expected, the held-out higher resolution ($112$) matches seen resolutions, since upsampling preserves digit topology. The drop at extreme downsampling ($7 \times 7$) reflects information loss---several digits become topologically ambiguous at this resolution---rather than discretization dependence.

\begin{figure}[t]
    \centering
    \includegraphics[width=\linewidth]{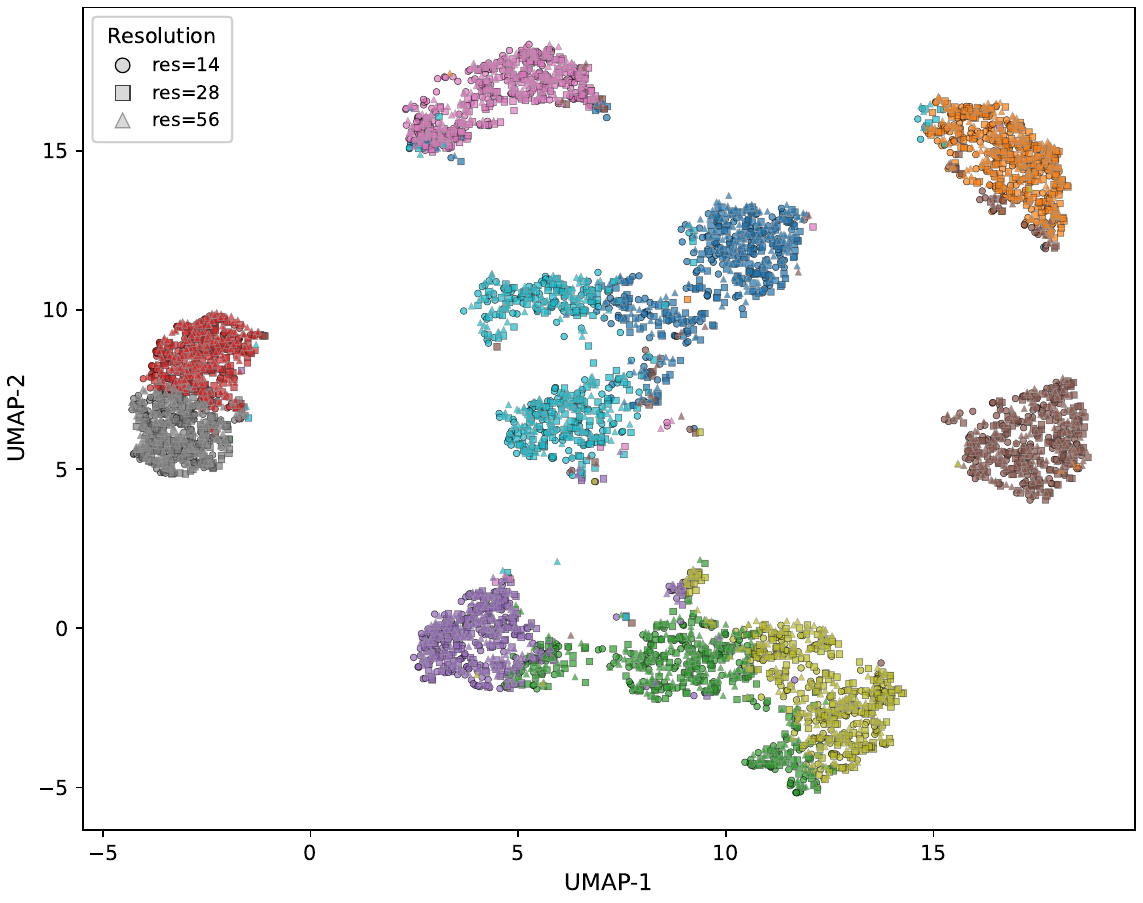}
    \caption{UMAP visualization of MNIST embedding. The marker shapes denote the different resolutions $r \in \{14, 28, 56\}$, and the points are colored by ground-truth digit label. Digit classes form coherent clusters overlapping across resolutions, indicating that the learned representation is primarily structured by semantic identity.}
    \label{fig:mnist_umap}
\end{figure}

Figure~\ref{fig:mnist_umap} visualizes the learned embeddings via UMAP. Digit classes form coherent clusters that overlap across resolutions, with no resolution-specific stratification. This is consistent with the hypernetwork capturing the structure of the underlying function rather than artifacts of the discretization grid.

\begin{table}[!h]
    \centering
    \caption{Clustering Performance (mean $\pm$ std over $5$ seeds) across discretization resolutions. Trained on multi-resolution $\mathrm{R_{train}} = \{14, 28, 56\}$; evaluated on seen and held-out resolutions $\mathrm{R_{test}} = \{7, 112\}$. ARI/AMI computed with K-Means (K=10). Stability across 14–56 and 112 indicates resolution invariance; degradation at 7 reflects severe undersampling.} 
    \label{tab:mnist_eval_diff_resolns}
    \begin{tabular}{c|c|c}
        \toprule 
        Resolution  & AMI $(\uparrow)$ & ARI $(\uparrow)$ \\
        \midrule 
        $14$ & $0.720 \pm 0.014$ & $0.604 \pm 0.035$\\
        $28$ & $0.724 \pm 0.013$ & $0.606 \pm 0.034$\\
        $56$ & $0.723 \pm 0.014$ & $0.607 \pm 0.034$\\
        \midrule
        $7$ & $0.578 \pm 0.008$ &  $0.456 \pm 0.034$\\
        $112$ & $0.723 \pm 0.014$ & $0.607 \pm 0.034$\\
        \bottomrule 
    \end{tabular}
\end{table}

\subsection{Kvasir dataset}
\label{subsec:medical_imaging}

\begin{figure}[t]
    \centering
    \setlength{\tabcolsep}{3pt}
    \renewcommand{\arraystretch}{1.1}

    \begin{tabular}{c *{3}{c}}
        & \textbf{Cluster 0} & \textbf{Cluster 1} & \textbf{Cluster 3} \\[0.4em]

        \rotatebox{90}{\large $64\times 64$} &
        \includegraphics[width=0.22\linewidth]{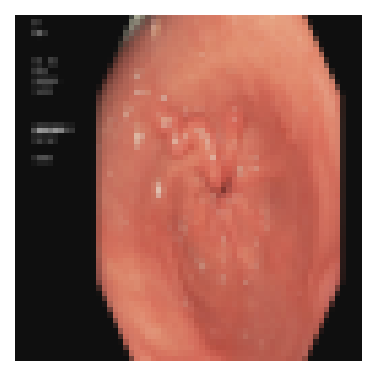} &
        \includegraphics[width=0.22\linewidth]{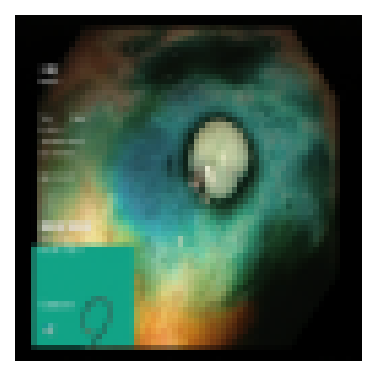} &
        \includegraphics[width=0.22\linewidth]{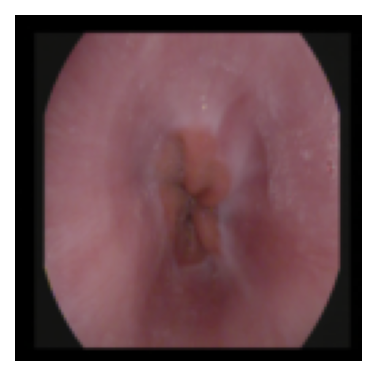} \\[0.6em]

        \rotatebox{90}{\large $128\times 128$} &
        \includegraphics[width=0.22\linewidth]{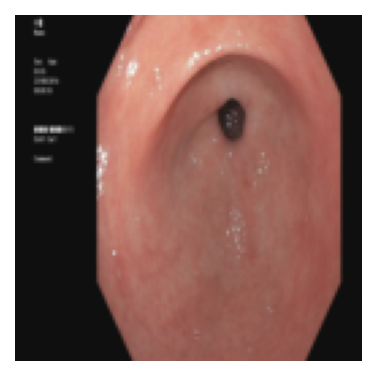} &
        \includegraphics[width=0.22\linewidth]{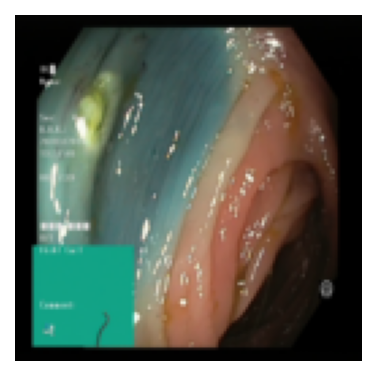} &
        \includegraphics[width=0.22\linewidth]{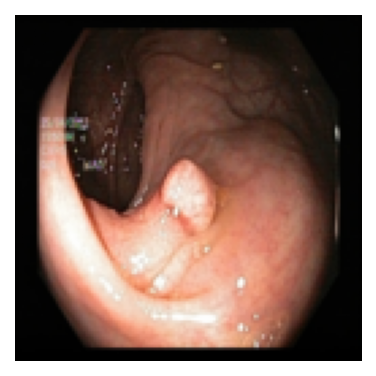} \\[0.6em]

        \rotatebox{90}{\large $256\times 256$} &
        \includegraphics[width=0.22\linewidth]{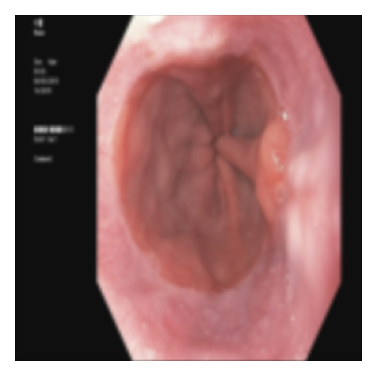} &
        \includegraphics[width=0.22\linewidth]{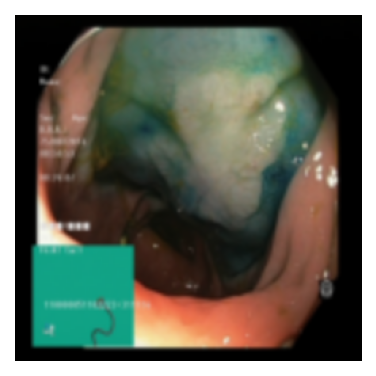} &
        \includegraphics[width=0.22\linewidth]{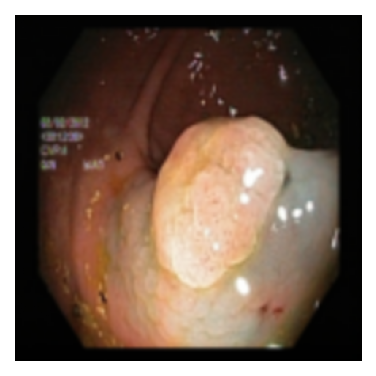} \\
    \end{tabular}

    \caption{Representative Kvasir samples arranged in a grid where rows correspond to discretization resolutions and columns correspond to learned clusters. Samples are mapped to their classes, irrespective of their resolution.}
    \label{fig:kvasir_cluster_samples}
    \vspace{-1.5em}
\end{figure}

Compared to MNIST, the Kvasir dataset~\cite{pogorelov2017kvasir}---consisting of eight semantic classes of gastrointestinal findings---presents substantially higher intra-class variability, complex texture, and real-world acquisition noise. Each RGB image is treated as a function $f : [0, 1]^2 \rightarrow \mathbb{R}^3$, mapping normalized spatial coordinates to RGB intensity values. Discretizations are constructed at resolutions $r \in \{64, 128, 256\}$ via bilinear interpolation, with training over $R_{\text{train}} = \{64, 128, 256\}$ and evaluation on held-out resolutions $R_{\text{test}} = \{32, 512\}$. The SIREN decoder has $d_z = 2{,}307$ parameters. Hypernetwork-representation learned clusters shown in Figure~\ref{fig:kvasir_cluster_samples} are clustered according to class identity, irrespective of sampling resolution.

\begin{figure}[t]
    \centering
    \includegraphics[width=\linewidth]{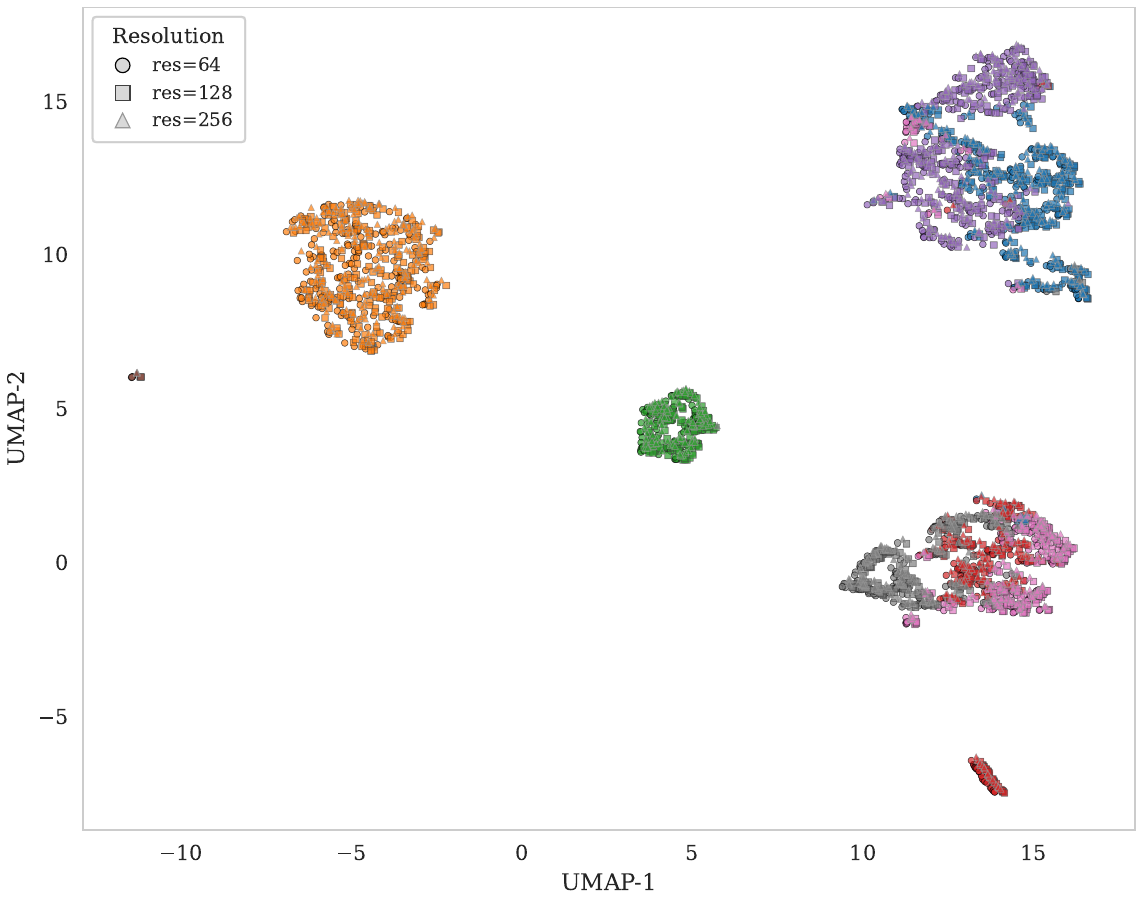}
    \caption{UMAP visualization of Kvasir embeddings. Ground-truth classes are represented by different colors and markers denote different resolutions. Compared to MNIST, class separation is less pronounced due to higher intra-class variability and visual ambiguity; however, embeddings from different resolutions overlap within each semantic group, indicating minimal resolution-induced stratification.}
    \label{fig:kvasir_umap}
\end{figure}

While absolute clustering scores are lower than in MNIST---reflecting the inherent difficulty of semantic grouping in medical imagery---Table~\ref{tab:kvasir_eval_diff_resolns} shows that performance remains statistically stable across all resolutions, including held-out ones. Figure~\ref{fig:kvasir_umap} confirms this observation: although clusters are less sharply separated than in MNIST, embeddings from different resolutions overlap within each semantic group, with no resolution-specific stratification. This indicates that the learned representation depends primarily on image content rather than grid resolution.

\begin{table}[!h]
    \centering
    \caption{Clustering Performance (mean $\pm$ std over $5$ seeds) across discretization resolutions. Trained on multi-resolution $\mathrm{R_{train}} = \{64, 128, 256\}$; evaluated on seen and held-out resolutions $\mathrm{R_{test}} = \{32, 512\}$. ARI/AMI computed with K-Means (K=$8$). AMI and ARI remain stable across discretizations.} 
    \label{tab:kvasir_eval_diff_resolns}
    \begin{tabular}{c|c|c}
        \toprule 
        Resolution  & AMI $(\uparrow)$ & ARI $(\uparrow)$ \\
        \midrule 
        $64$ & $0.479 \pm 0.014$ & $0.311 \pm 0.010$ \\
        $128$ & $0.481 \pm 0.013$ & $0.312 \pm 0.009$\\
        $256$ & $0.481 \pm 0.015$ & $0.313 \pm 0.010$ \\
        \midrule
        $32$ & $0.479 \pm 0.014$ & $0.311 \pm 0.010$ \\
        $512$ & $0.481 \pm 0.014$ & $0.311 \pm 0.010$ \\
        \bottomrule 
    \end{tabular}
\end{table}

\subsection{ERing dataset}
\label{subsec:time_series}

The ERing multivariate dataset from the UEA archive~\citep{bagnall2018uea} captures finger gestures recorded by a wearable ring equipped with electric field sensors. We select this dataset specifically because its univariate input domain ($\Omega \subset \mathbb{R}$) falls within the scope of FAEclust~\citep{singh2025faeclust}---the closest baseline---making it the one setting where a direct comparison is meaningful. Each time-series sample is treated as a function $f\colon [0, 1] \rightarrow \mathbb{R}^4$, mapping normalized time to the four sensor channels. Temporal discretizations are constructed at $T \in \{33, 65, 130\}$ by uniformly resampling the time axis, with training over $T_{\text{train}} = \{33, 65, 130\}$ and evaluation on held-out resolutions $T_{\text{test}} = \{16, 260\}$. The SIREN decoder has $d_z = 94$ parameters.

\begin{figure*}[t]
    \centering
    \begin{minipage}[c]{0.38\linewidth}
        \centering
        \includegraphics[width=\linewidth]{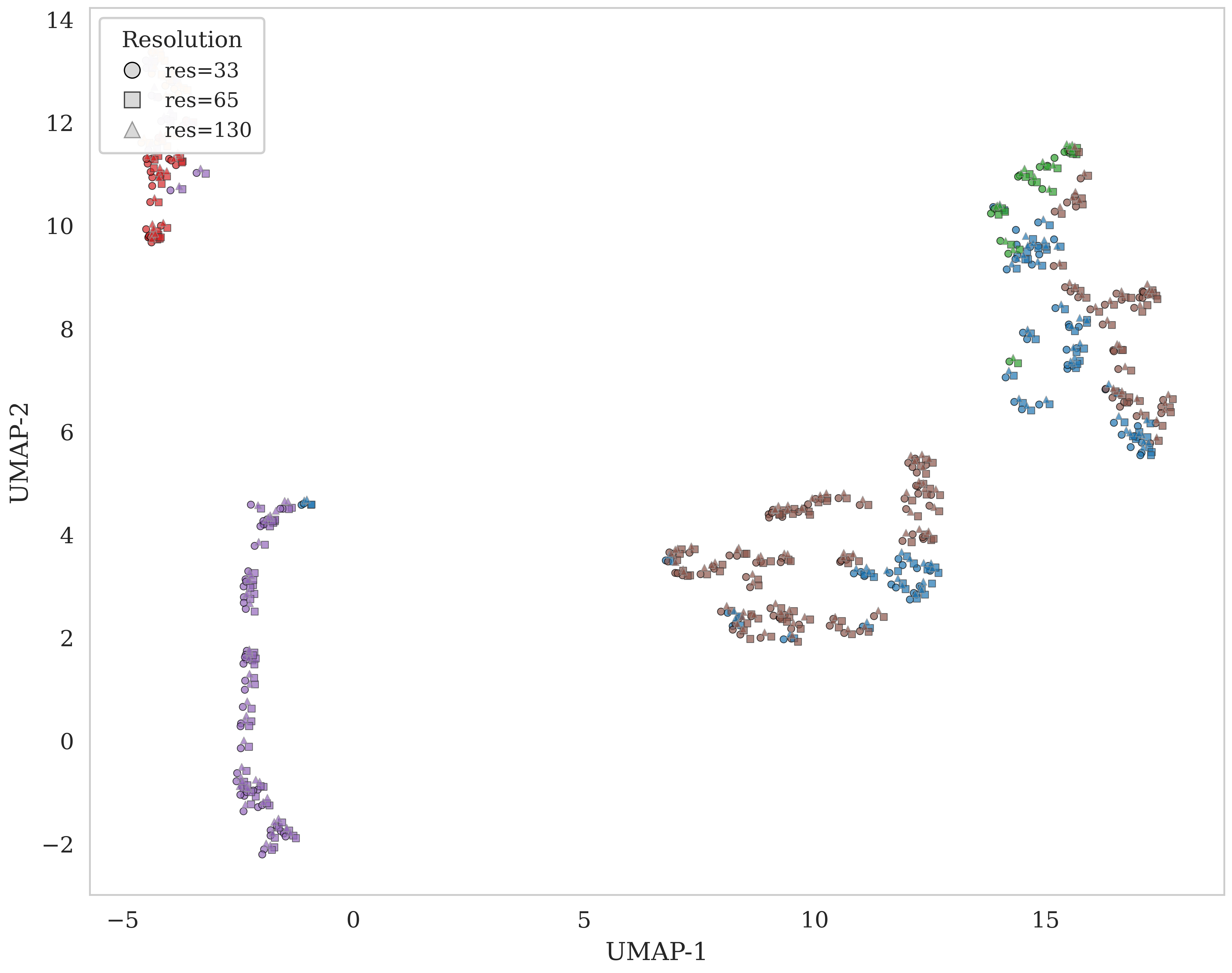}
    \end{minipage}%
    \hfill
    \begin{minipage}[c]{0.59\linewidth}
        \centering
        \setlength{\tabcolsep}{4pt}
        \renewcommand{\arraystretch}{1.1}
        \begin{tabular}{c *{3}{c}}
            & \textbf{Cluster 0} & \textbf{Cluster 1} & \textbf{Cluster 3} \\[0.4em]

            \rotatebox{90}{\small $\mathrm{t = 33}$} &
            \includegraphics[width=0.29\linewidth]{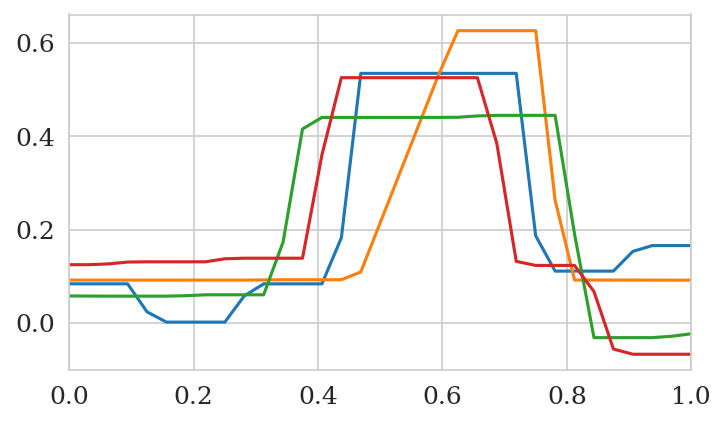} &
            \includegraphics[width=0.29\linewidth]{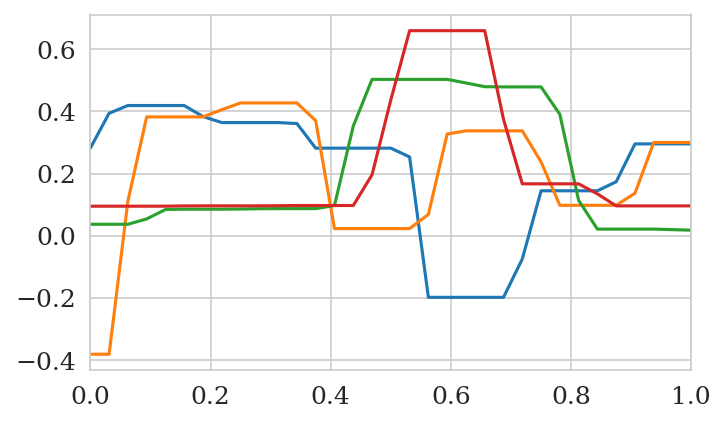} &
            \includegraphics[width=0.29\linewidth]{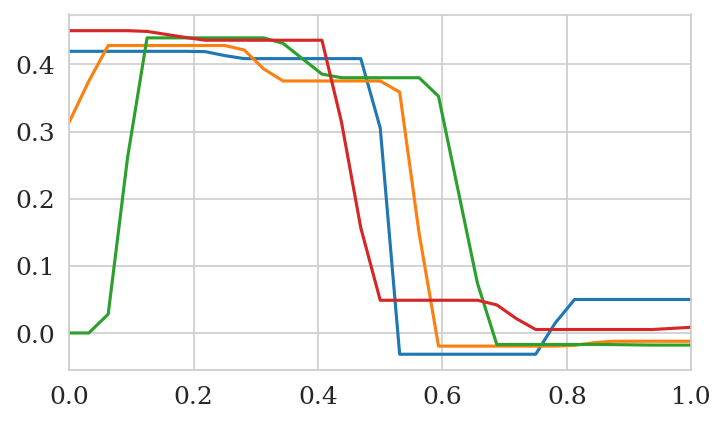} \\[0.4em]

            \rotatebox{90}{\small $\mathrm{t = 65}$} &
            \includegraphics[width=0.29\linewidth]{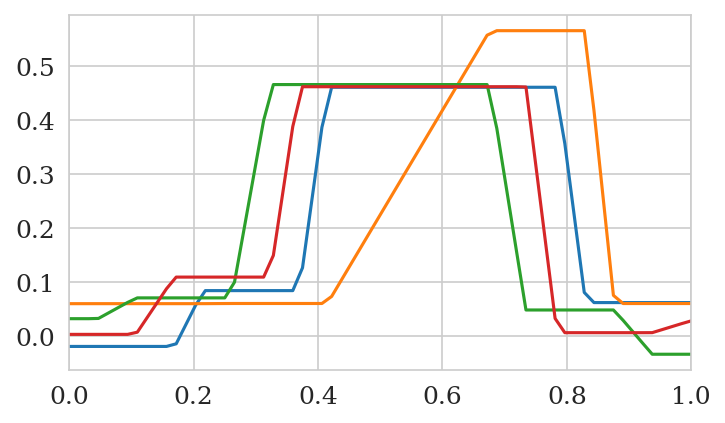} &
            \includegraphics[width=0.29\linewidth]{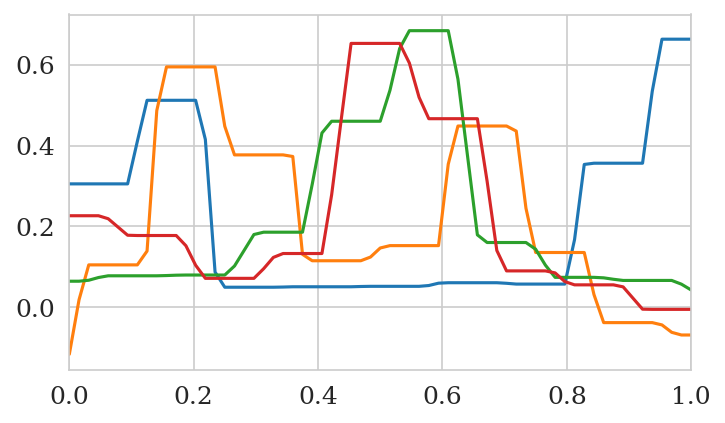} &
            \includegraphics[width=0.29\linewidth]{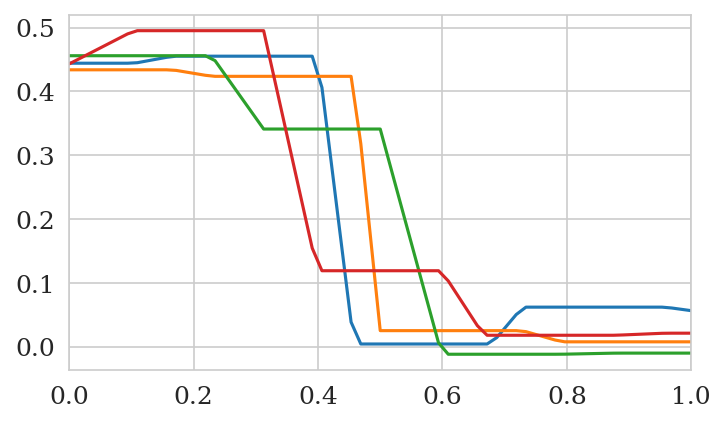} \\[0.4em]

            \rotatebox{90}{\small $\mathrm{t = 130}$} &
            \includegraphics[width=0.29\linewidth]{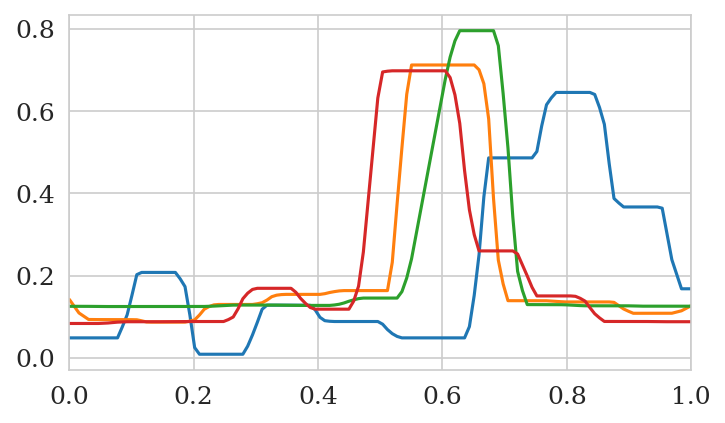} &
            \includegraphics[width=0.29\linewidth]{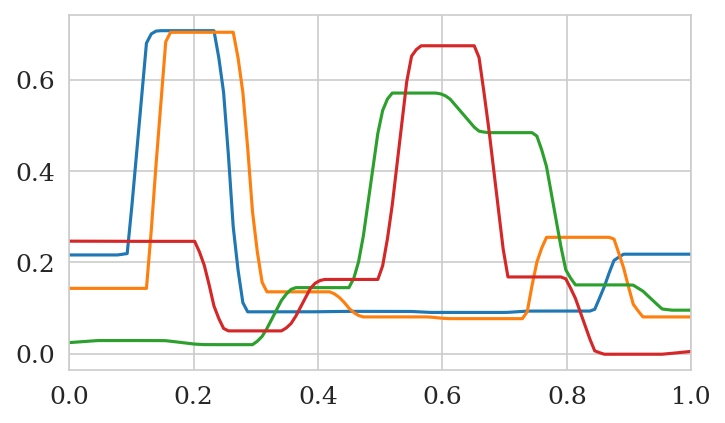} &
            \includegraphics[width=0.29\linewidth]{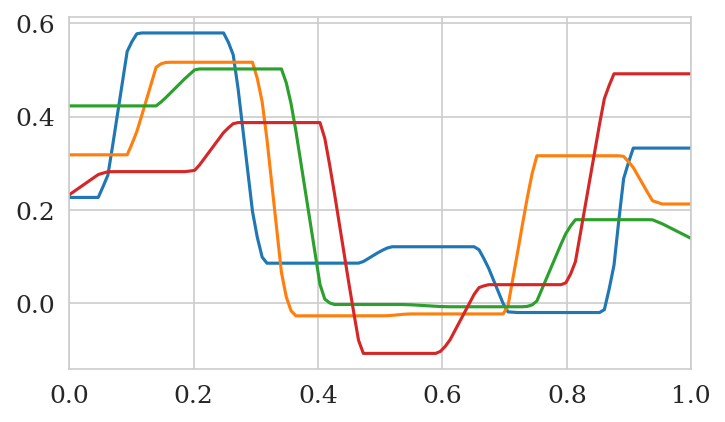} \\
        \end{tabular}
    \end{minipage}
    \caption{ERing embeddings and cluster samples. \textbf{Left:} UMAP visualization; while class separability is less than in MNIST, embeddings do not stratify by timesteps. \textbf{Right:} Representative samples arranged by discretization timestep (rows) and learned cluster (columns). Class identity is preserved irrespective of discretization.}
    \label{fig:ering_umap}
    \label{fig:ering_cluster_samples}
    \vspace{-1.5em}
\end{figure*}

Table~\ref{tab:ering_eval_diff_resolns} shows that performance remains stable across both seen and held-out discretizations, with AMI~$\approx 0.56$. FAEclust, which jointly optimizes a clustering-specific loss and is designed specifically for functional time-series data, reports AMI~$= 0.664$ on ERing~\citep{singh2025faeclust}. While our method does not match this score, it is worth noting that our framework trains a pure reconstruction objective without any clustering supervision, and the same framework handles MNIST and Kvasir---settings where FAEclust is inapplicable due to its restriction to one-dimensional input domains. As observed in Figure~\ref{fig:ering_cluster_samples}, clustering preserves class identity irrespective of temporal discretization. Figure~\ref{fig:ering_umap} confirms that embeddings do not stratify by resolution despite moderate class separation, indicating that the learned representation is invariant to temporal discretization.

\begin{table}[!h]
    \centering
    \caption{Clustering Performance (mean $\pm$ std over $5$ seeds) across discretization timesteps. Trained on multi-resolution $\mathrm{R_{train}} = \{33, 65, 130\}$; evaluated on seen and held-out resolutions $\mathrm{R_{test}} = \{16, 260\}$. ARI/AMI computed with K-Means (K=$6$). AMI and ARI remain unaffected across different discretizations.} 
    \label{tab:ering_eval_diff_resolns}
    \begin{tabular}{c|c|c}
        \toprule 
        Resolution & AMI $(\uparrow)$ & ARI $(\uparrow)$ \\
        \midrule 
        $33$ & $0.564 \pm 0.041$ & $0.408 \pm 0.049$ \\
        $65$ & $0.559 \pm 0.042$ & $0.404 \pm 0.051$\\
        $130$ & $0.559 \pm 0.042$ & $0.403 \pm 0.049$ \\
        \midrule
        $16$ & $0.558 \pm 0.049$ & $0.404 \pm 0.056$ \\
        $260$ & $0.556 \pm 0.044$ & $0.402 \pm 0.051$ \\
        \bottomrule 
    \end{tabular}
\end{table}

\paragraph{Clustering algorithm comparison.} To verify that the learned representation is not tied to a particular clustering algorithm, we replace $K$-means with a Gaussian mixture model (GMM) and report results in Appendices~\ref{appendix:mnist},~\ref{appendix:medical}, and~\ref{appendix:ering}. Across all three datasets, GMM yields performance comparable to $K$-means, confirming that the weight-space embeddings carry cluster structure that can be exploited by different off-the-shelf algorithms without retraining.

\section{Related work}
\label{sec:related}

Classical functional data clustering relies on basis expansions such as B-splines, wavelets, or functional PCA, followed by clustering in coefficient space \citep{abraham2003bspline, tarpey2003clustering, ramsay2005functional, giacofci2013wavelet, jacques2014survey}. Model-based approaches cluster functional PCA scores under Gaussian mixture assumptions \citep{chiou2007functional, bouveyron2011funhddc, jacques2013funclust}, while \citet{james2003clustering} handle sparse and irregular observations through mixed-effects spline models---but all remain restricted to univariate domains and a fixed basis system. Recent work replaces handcrafted bases with learned representations: functional multilayer perceptrons \citep{rossi2002functional}, functional autoencoders (FAE) \citep{hsieh2021functional}, and variational autoencoding neural operators (VANO) \citep{seidman2023vano} learn latent embeddings of discretized functions. However, these methods operate on functions evaluated on a common grid and therefore remain discretization-dependent.

The closest prior work is FAEclust \citep{singh2025faeclust}, which jointly trains a functional autoencoder with a clustering objective. However, its integral-operator architecture restricts the input domain to $\mathcal{T} \subset \mathbb{R}$, requires a shared grid, and couples the representation to a specific clustering criterion. A direct comparison is therefore limited to the ERing experiment (Section~\ref{subsec:time_series}), where FAEclust achieves higher AMI but relies on clustering-specific supervision that our reconstruction-only framework forgoes.

In contrast, our approach encodes raw coordinate--value pairs via a mesh-independent aggregation mechanism \citep{stuart2025autoencoders}, predicts INR weights \citep{sitzmann2020siren} using a hypernetwork \citep{ha2017hypernetworks, mayer2024ple}, and decouples representation learning from clustering under multi-resolution training \citep{sakarvadia2026false}. This deliberate decoupling means that any off-the-shelf clustering algorithm can be applied in the learned weight space without retraining, yielding a framework that is simultaneously grid-independent and clustering-algorithm-agnostic.

\section{Discussion and future work}
\label{sec:discussion}

The central finding across image, medical, and multivariate time-series experiments is not merely competitive clustering accuracy, but \emph{resolution invariance}: by mapping functions into INR weight space, cluster assignments become decoupled from the discretization at which observations are recorded. This invariance holds even at resolutions absent from training, breaking down only under extreme information loss---e.g., MNIST at $7 \times 7$, where the underlying signal content itself becomes indistinguishable across classes. The framework thus reframes functional data clustering as a problem of geometry in neural weight space rather than in discretized observation space.

By design, the framework deliberately separates representation learning from clustering: the hypernetwork is trained solely via reconstruction, and clustering is performed post hoc in weight space. While this decoupling enables algorithmic flexibility---practitioners can swap clustering algorithms without retraining---it does not explicitly optimize cluster separation. Methods that jointly train a clustering objective, such as FAEclust \citep{singh2025faeclust}, may achieve tighter within-cluster compactness on their target metric, but at the cost of tying the representation to a single algorithm and a fixed discretization.

From a practical standpoint, when the output dimensionality $p$ of the target function is very large, the INR's final layer scales as $\mathcal{O}(h \cdot p)$---where $h$ is the hidden width---inflating the weight vector and potentially hindering clustering in weight space. In practice, most scientifically relevant functional datasets are characterized by dense spatial or temporal grids but modest per-point feature dimensions (e.g., scalar, RGB, or three-component vector fields), so this limitation rarely arises.

Several directions remain open. First, the geometry of the induced weight space---including its curvature, connectivity, and the extent to which functionally similar inputs map to metrically nearby weights---is not yet well understood and warrants theoretical investigation. Second, incorporating clustering-aware objectives into the reconstruction loss may improve cluster separability while preserving the framework's generality. Third, because the hypernetwork amortizes the mapping from observations to weight space, new functions can be embedded via a single forward pass without retraining; this opens the door to streaming or online clustering settings in which data arrive sequentially and cluster assignments must be updated incrementally. Finally, replacing the SIREN decoder with wavelet-based INRs~\citep{roddenberry2024inr} could yield weight-space representations that better capture spatially localized features---e.g., edges or transients---since wavelets simultaneously localize in frequency and space, unlike the global sinusoidal activations used by SIREN.

\begin{acknowledgements}
    Ali Siahkoohi acknowledges support from the Institute for Artificial Intelligence at the University of Central Florida.
\end{acknowledgements}

\bibliography{uai2026-template}

\clearpage

\appendix

\section{MNIST Dataset}
\label{appendix:mnist}

\begin{figure}[!h]
    \centering
    \includegraphics[width=0.95\linewidth]{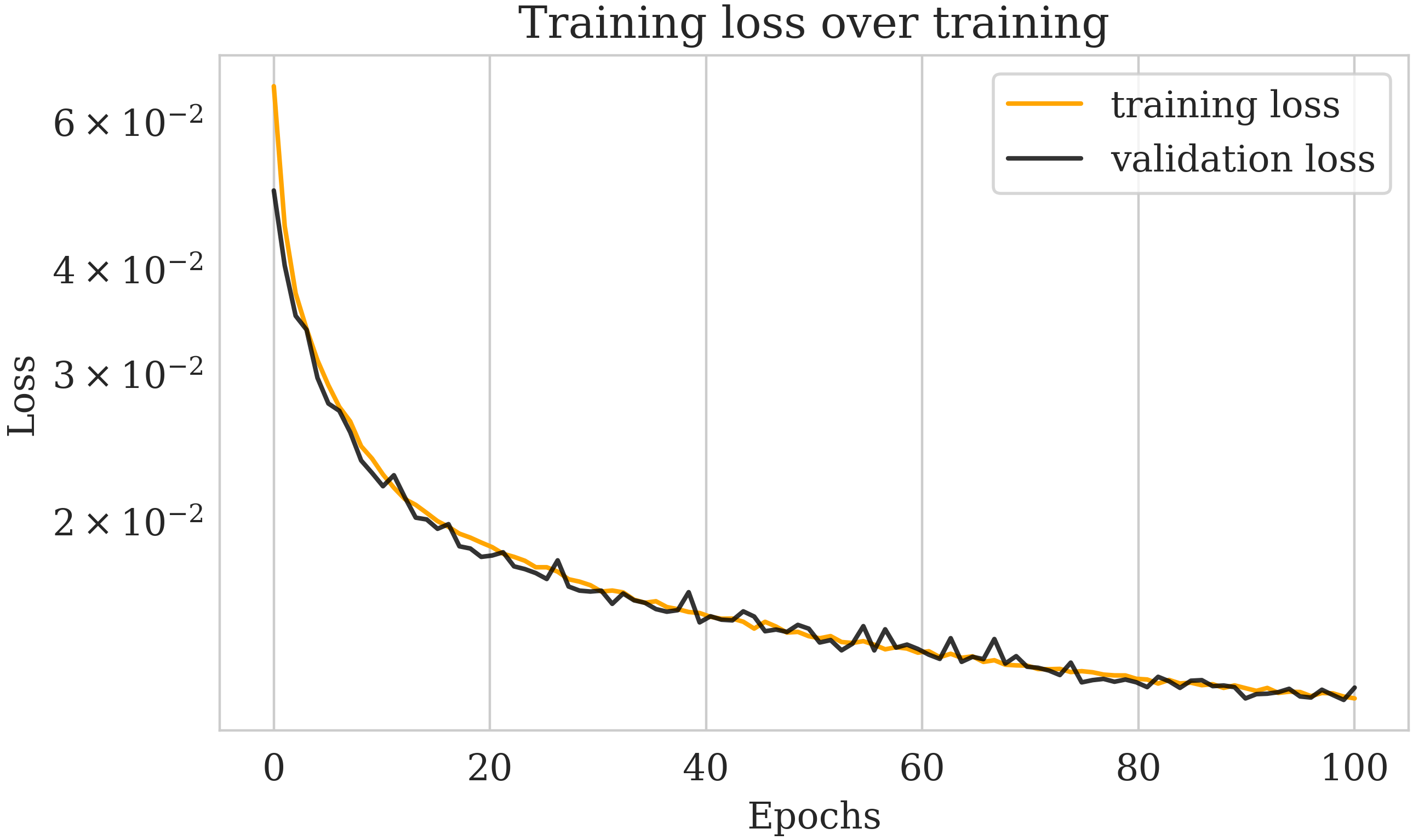}
    \caption{Training and validation loss curves for MNIST.}
    \label{fig:mnist_loss_curves}
\end{figure}

Additional results and visualizations for the MNIST experiments are presented below. Figure~\ref{fig:mnist_clusters_results} shows representative members from four learned clusters across resolutions, confirming that cluster membership is determined by digit identity rather than discretization. Figure~\ref{fig:mnist_loss_curves} reports training and validation loss curves. Table~\ref{tab:mnist_cmp_clust} compares $K$-means and GMM clustering in the learned weight space, demonstrating that the representation supports multiple clustering algorithms without retraining.

\begin{figure}[ht]
    \centering

    \begin{subfigure}[t]{0.48\linewidth}
        \centering
        \includegraphics[width=\linewidth]{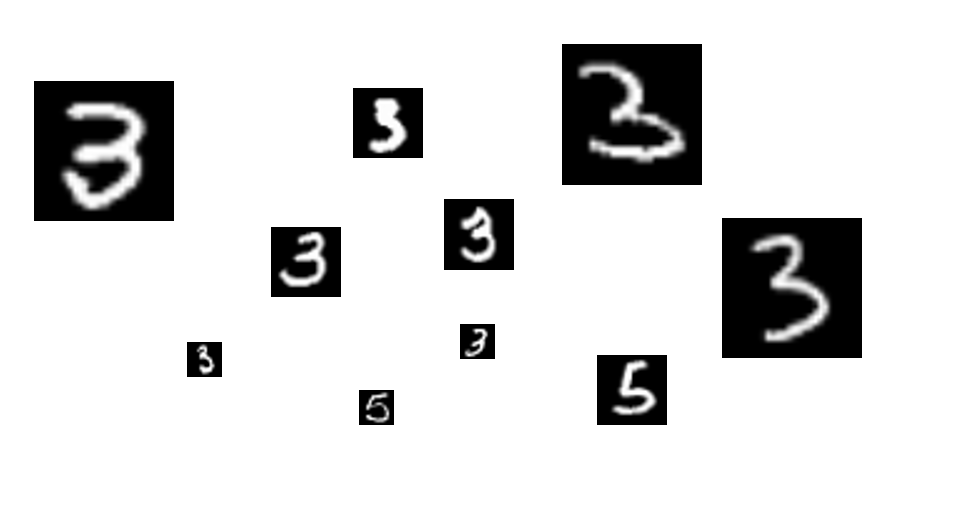}
        \caption{Cluster $3$}
        \label{subfig:mnist_cluster_3}
    \end{subfigure}\hfill
    \begin{subfigure}[t]{0.48\linewidth}
        \centering
        \includegraphics[width=\linewidth]{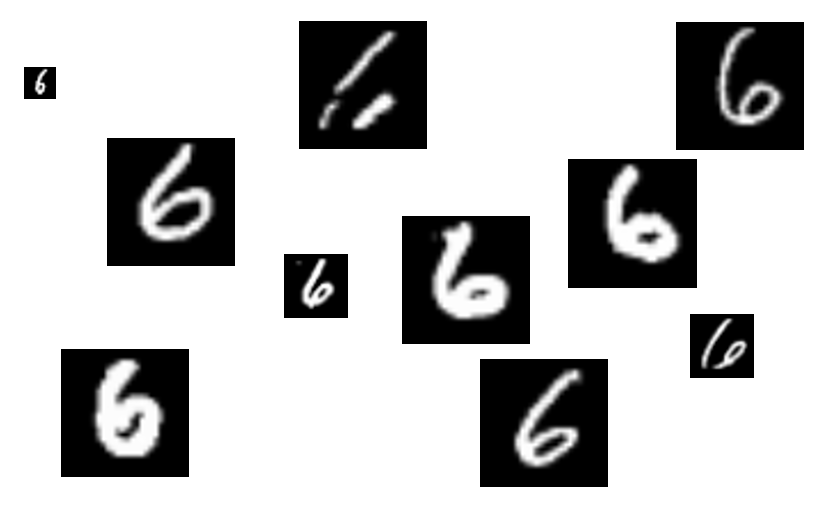}
        \caption{Cluster $6$}
        \label{subfig:mnist_cluster_6}
    \end{subfigure}
    \vspace{0.6em}

    \begin{subfigure}[t]{0.48\linewidth}
        \centering
        \includegraphics[width=\linewidth]{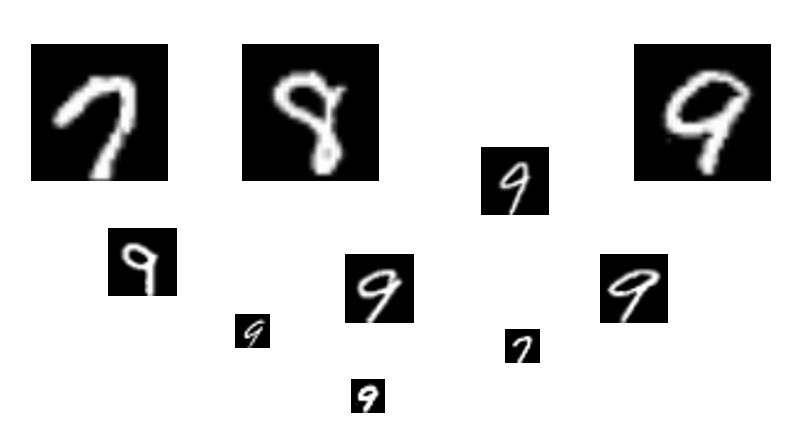}
        \caption{Cluster $9$}
        \label{subfig:mnist_cluster_9}
    \end{subfigure}\hfill
    \begin{subfigure}[t]{0.48\linewidth}
        \centering
        \includegraphics[width=\linewidth]{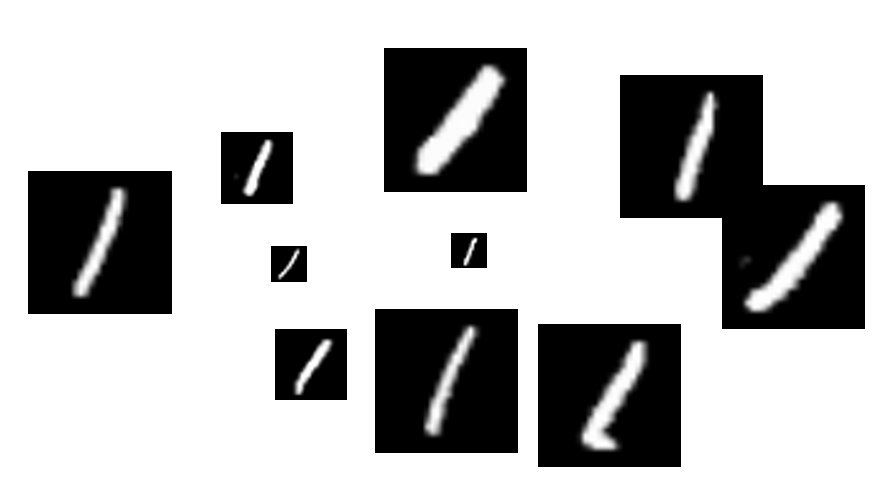}
        \caption{Cluster $1$}
        \label{subfig:mnist_cluster_1}
    \end{subfigure}

    \caption{Representative members from four learned clusters on MNIST. The learned embeddings maintain class separation and avoid resolution-level stratification.}
    \label{fig:mnist_clusters_results}
    \vspace{-1.5em}
\end{figure}

\begin{table*}[!h]
    \centering
    \caption{Clustering algorithm comparison on MNIST. Trained on $\mathrm{R_{train}} = \{14, 28, 56\}$; evaluated on seen and held-out resolutions $\mathrm{R_{test}} = \{7, 112\}$. $K = 10$ for both $K$-means and GMM.}
    \label{tab:mnist_cmp_clust}
    \begin{tabular}{c|c|c|c|c}
        \toprule
        Resolution  & KMeans(AMI) $(\uparrow)$ & KMeans(ARI) $(\uparrow)$ & GMM(AMI) $(\uparrow)$ & GMM(ARI) $(\uparrow)$\\
        \midrule
        $14$ & $0.713$ & $0.586$ & $0.703$ & $0.589$ \\
        $28$ & $0.717$ & $0.589$ & $0.702$ & $0.590$ \\
        $56$ & $0.750$ & $0.674$ & $0.706$ & $0.597$ \\
        \midrule
        $7$ & $0.593$ & $0.525$ & $0.414$ & $0.149$ \\
        $112$ & $0.716$ & $0.590$ & $0.701$ & $0.589$ \\
        \bottomrule
    \end{tabular}
\end{table*}

\section{Kvasir dataset}
\label{appendix:medical}

Additional results and visualizations for the Kvasir experiments are presented below. Figure~\ref{fig:kvasir_loss_curves} reports training and validation loss curves over $100$ epochs; the training loss decreases steadily and the gap between training and validation losses remains small, indicating that the model generalizes without substantial overfitting. Table~\ref{tab:kvasir_cmp_clust} compares $K$-means and GMM clustering in the learned weight space. Both algorithms yield comparable AMI and ARI scores across all resolutions---including held-out ones---confirming that the representation carries intrinsic cluster structure that is not tied to a particular clustering method.

\begin{figure}[!h]
    \centering
    \includegraphics[width=0.95\linewidth]{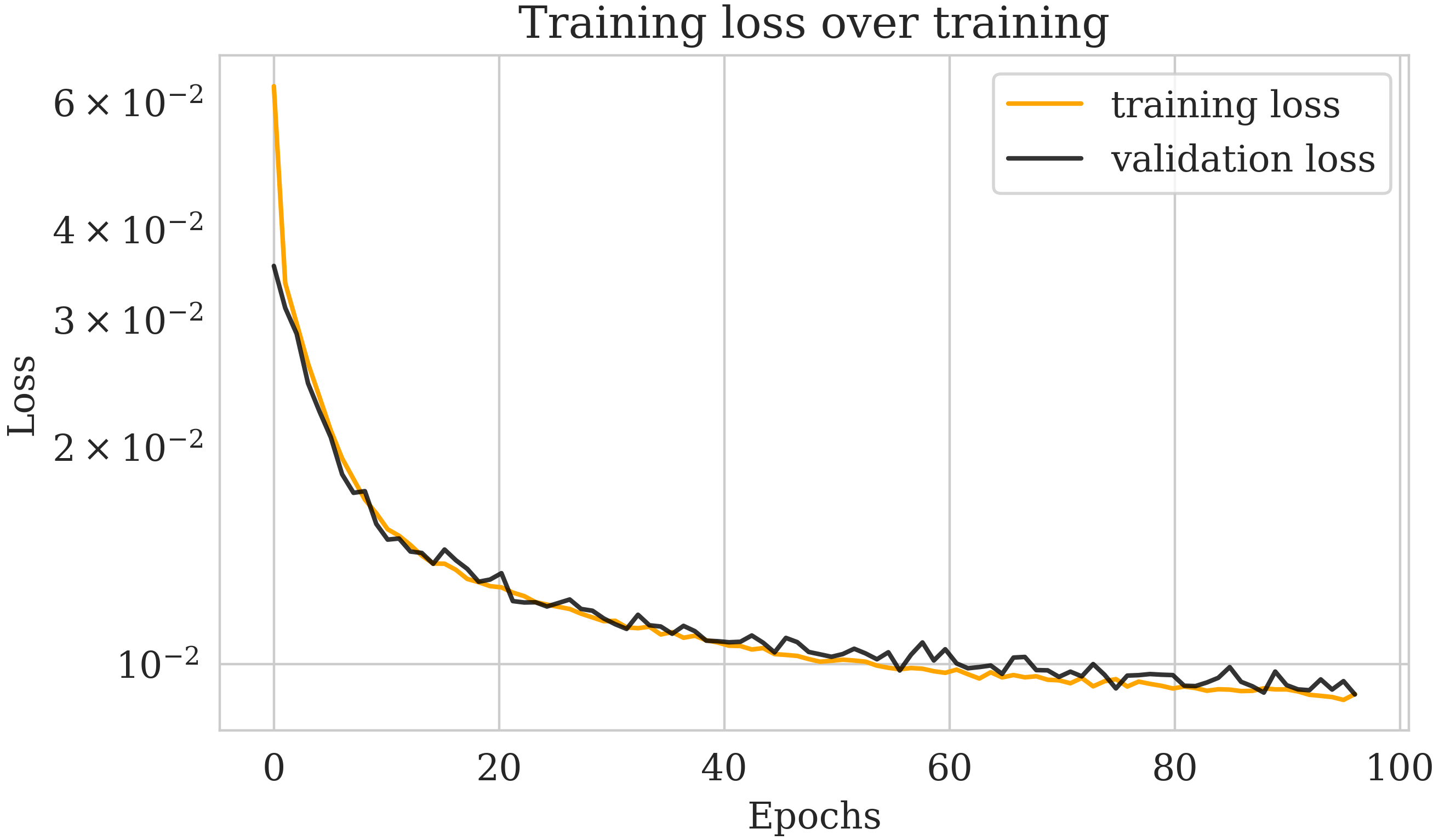}
    \caption{Training and validation loss curves for the Kvasir dataset.}
    \label{fig:kvasir_loss_curves}
\end{figure}

\begin{table*}[!h]
    \centering
    \caption{Clustering algorithm comparison on Kvasir. Trained on $\mathrm{R_{train}} = \{64, 128, 256\}$; evaluated on seen and held-out resolutions $\mathrm{R_{test}} = \{32, 512\}$. $K = 8$ for both $K$-means and GMM.}
    \label{tab:kvasir_cmp_clust}
    \begin{tabular}{c|c|c|c|c}
        \toprule
        Resolution  & KMeans(AMI) $(\uparrow)$ & KMeans(ARI) $(\uparrow)$ & GMM(AMI) $(\uparrow)$ & GMM(ARI) $(\uparrow)$\\
        \midrule
        $64$ & $0.492$ & $0.311$ & $0.467$ & $0.320$ \\
        $128$ & $0.493$ & $0.314$ & $0.472$ & $0.295$ \\
        $256$ & $0.499$ & $0.320$ & $0.472$ & $0.295$ \\
        \midrule
        $32$ & $0.493$ & $0.313$ & $0.474$ & $0.299$ \\
        $512$ & $0.494$ & $0.313$ & $0.472$ & $0.295$ \\
        \bottomrule
    \end{tabular}
\end{table*}

\section{ERing dataset}
\label{appendix:ering}

\begin{figure}[!h]
    \centering
    \includegraphics[width=0.95\linewidth]{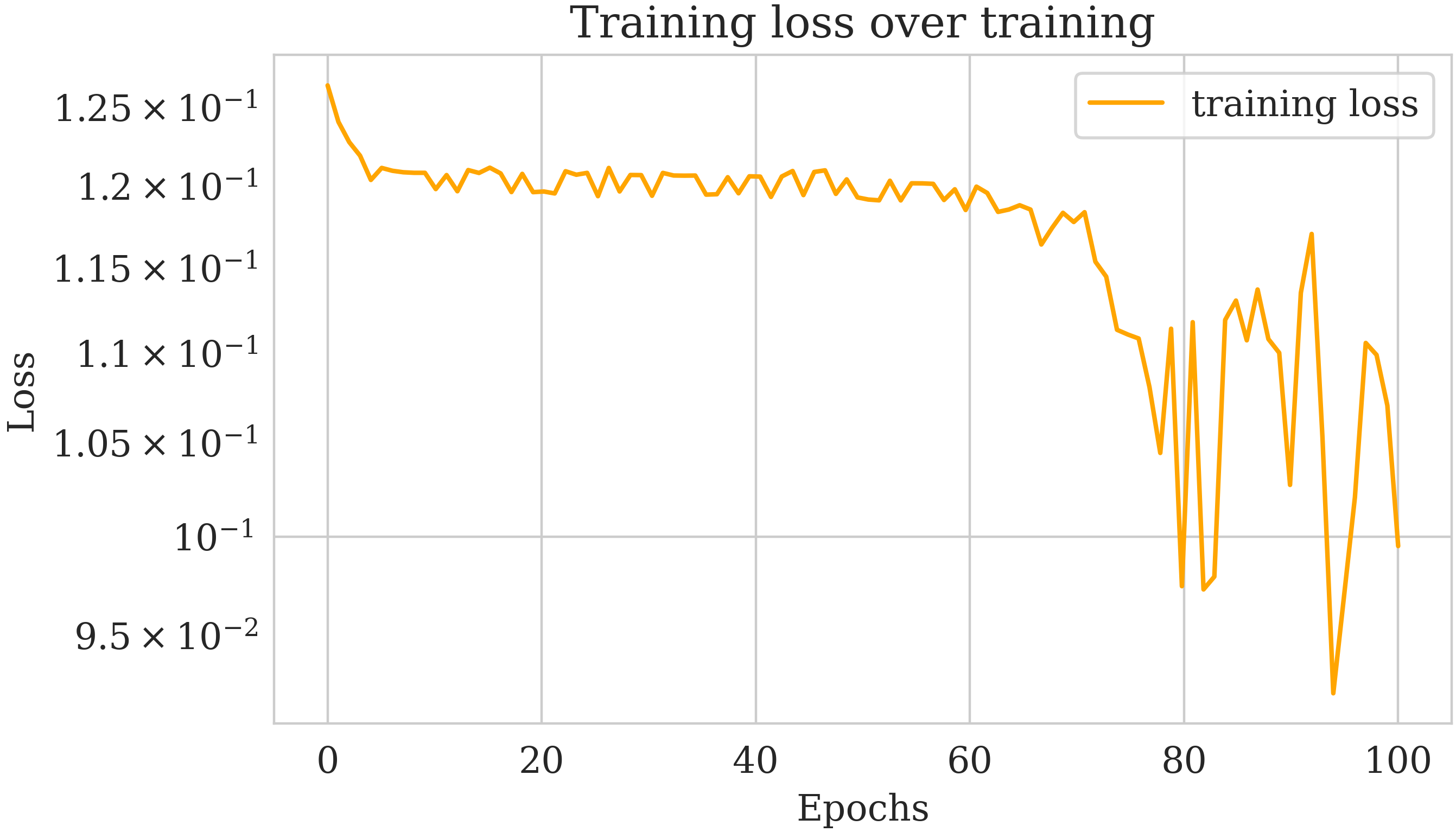}
    \caption{Training loss curve for the ERing dataset.}
    \label{fig:ering_loss_curves}
\end{figure}

Additional results and visualizations for the ERing experiments are presented below. The ERing dataset contains only $30$ training samples, making it a challenging setting for representation learning. Figure~\ref{fig:ering_loss_curves} reports the training loss curve, which converges smoothly. Table~\ref{tab:ering_cmp_clust} compares $K$-means and GMM clustering in the learned weight space. Both algorithms produce similar AMI scores across all temporal resolutions, including held-out discretizations ($T = 16$ and $T = 260$), confirming that the learned representation is compatible with different clustering methods even in the small-sample regime.

\begin{table*}[!h]
    \centering
    \caption{Clustering algorithm comparison on ERing. Trained on $\mathrm{R_{train}} = \{33, 65, 130\}$; evaluated on seen and held-out resolutions $\mathrm{R_{test}} = \{16, 260\}$. $K = 6$ for both $K$-means and GMM.}
    \label{tab:ering_cmp_clust}
    \begin{tabular}{c|c|c|c|c}
        \toprule
        Resolution  & KMeans(AMI) $(\uparrow)$ & KMeans (ARI) $(\uparrow)$ & GMM (AMI) $(\uparrow)$ & GMM (ARI) $(\uparrow)$\\
        \midrule
        $33$ & $0.564$ & $0.439$ & $0.567$ & $0.373$ \\
        $65$ & $0.556$ & $0.432$ & $0.553$ & $0.365$ \\
        $130$ & $0.553$ & $0.428$ & $0.553$ & $0.365$ \\
        \midrule
        $16$ & $0.553$ & $0.424$ & $0.584$ & $0.388$ \\
        $260$ & $0.553$ & $0.166$ & $0.552$ & $0.363$ \\
        \bottomrule
    \end{tabular}
\end{table*}

\section{Implementation Details}
\label{appendix:impl_details}

Across all experiments we use the same hypernetwork design, although per-dataset configurations (SIREN width, batch size, number of epochs) differ; see Table~\ref{tab:hyperparams} for details. Each input function is represented as an unordered set of $I_n$ coordinate-value pairs $\{(\bm{x}_i, u(\bm{x}_i))\}_{i=1}^{I_n}$, following the notation in Section~\ref{sec:problem-def}. The coordinates $\bm{x}_i$ are embedded with random Fourier features (RFF)~\citep{rahimi2007random} of dimension $d_{\mathrm{rff}} = 32$. The per-point network $h^{(1)}$ is a $3$-layer multi-layer perceptron (MLP) with hidden width $64$ that processes each $(\mathrm{RFF}(\bm{x}_i),\, u(\bm{x}_i))$ pair independently; mean pooling aggregates these into a $64$-dimensional representation. The weight predictor $h^{(2)}$ maps this representation to the full parameter vector $\bm{w}$ of a $4$-layer SIREN decoder; no per-instance INR optimization is performed. The training objective is the mean squared reconstruction error in equation~\eqref{eq:loss}.

\begin{table}[h]
\centering
\caption{Per-dataset hyperparameter settings.}
\label{tab:hyperparams}
\small
\begin{tabular}{lccc}
\toprule
 & MNIST & Kvasir & ERing \\
\midrule
SIREN hidden width & 5 & 32 & 5 \\
SIREN layers & 4 & 4 & 4 \\
$d_z$ (SIREN params) & 81 & 2{,}307 & 94 \\
Batch size & 128 & 64 & 16 \\
Epochs & 500 & 100 & 100 \\
$\omega_0$ & 30 & 30 & 30 \\
LR & $3\text{e-}4 \!\to\! 1\text{e-}4$ & $3\text{e-}4 \!\to\! 1\text{e-}4$ & $3\text{e-}4 \!\to\! 1\text{e-}4$ \\
Coord.\ dim $d$ & 2 & 2 & 1 \\
Output dim $m$ & 1 & 3 & 4 \\
\bottomrule
\end{tabular}
\end{table}

Multi-resolution training is enabled by randomly generating a resolution $r \sim \mathcal{U}(\mathrm{R_{train}})$ at each iteration, and held-out resolutions $\mathrm{R_{test}}$ are used to assess discretization invariance. All models are optimized with Adam~\citep{kingma2015adam} ($\beta_1 = 0.9$, $\beta_2 = 0.999$) using a power-law learning rate decay from $3 \times 10^{-4}$ to $1 \times 10^{-4}$, with no weight decay. Per-dataset batch sizes and epoch counts are listed in Table~\ref{tab:hyperparams}. Results are reported over $5$ random seeds as mean $\pm$ std. Unless stated otherwise, we use an "oracle-K" protocol with $K$ equal to the number of ground-truth classes.

\end{document}